\documentclass{article}
\usepackage{spconf,amsmath,graphicx}

\usepackage{caption}
\usepackage{subcaption}
\usepackage{graphicx,wrapfig,lipsum}
\usepackage{url} 
\usepackage{hyperref}
\usepackage{booktabs} 
\usepackage{multirow}
\usepackage{comment}

\title{Let's Roll: Synthetic Dataset Analysis for Pedestrian Detection Across Different Shutter Types\thanks{T\lowercase{his research is partially supported by a grant from} S\lowercase{amsung}.}}
\name{Yue Hu$^1$, Gourav Datta$^1$, Kira Beerel$^2$, Peter Beerel$^1$}
\address{$^1$University of Southern California \quad $^2$ Harvard Westlake High School}
\begin{document}
%
\maketitle
\begin{abstract}
    
Computer vision (CV) pipelines are typically evaluated on datasets processed by image signal processing (ISP) pipelines even though, for resource-constrained applications, an important research goal is to avoid as many ISP steps as possible. In particular, most CV datasets consist of global shutter (GS) images even though most cameras today use a rolling shutter (RS). This paper studies the impact of different shutter mechanisms on machine learning (ML) object detection models on a synthetic dataset that we generate using the advanced simulation capabilities of Unreal Engine 5 (UE5). 
In particular, we train and evaluate mainstream detection models with our synthetically-generated paired GS and RS datasets to ascertain whether there exists a significant difference in detection accuracy between these two shutter modalities, especially when capturing low-speed objects (e.g., pedestrians). 
The results of this emulation framework indicate the performance between them are remarkably congruent for coarse-grained detection (mean average precision (mAP) for IOU=0.5), but have significant differences for fine-grained measures of detection accuracy (mAP for IOU=0.5:0.95). This implies that ML pipelines might not need explicit correction for RS for many 
object detection applications, but mitigating RS effects in ISP-less ML pipelines that target fine-grained location of the objects 
may need additional research.

\end{abstract}
\section{Introduction}

In the field of digital photography and videography, the choice of camera shutter mechanism plays a pivotal role in determining the quality and fidelity of captured images. The majority of mainstream cameras available in the market today employ a RS mechanism~\cite{dai2016rolling,fan2021sunet}. While this mechanism offers several advantages, including reduced manufacturing costs and lower power consumption, it often distorts the image, particularly when there is relative motion between the camera and the subject being captured. Algorithms within image signal processing (ISP) pipelines correct for this distortion~\cite{lao2018robust,rengarajan2017unrolling,zhuang2019learning,liu2020deep,zhuang2017rolling,cao2022learning}. By doing so, the resulting images, when viewed by the human eye, appear undistorted and true to the scene. These corrected images are typically also used as inputs to a CV pipeline, ensuring that the models are not adversely affected by the distortions inherent in RS captures. 




\begin{figure}[!t]
    \centering
    \includegraphics[width=\linewidth]{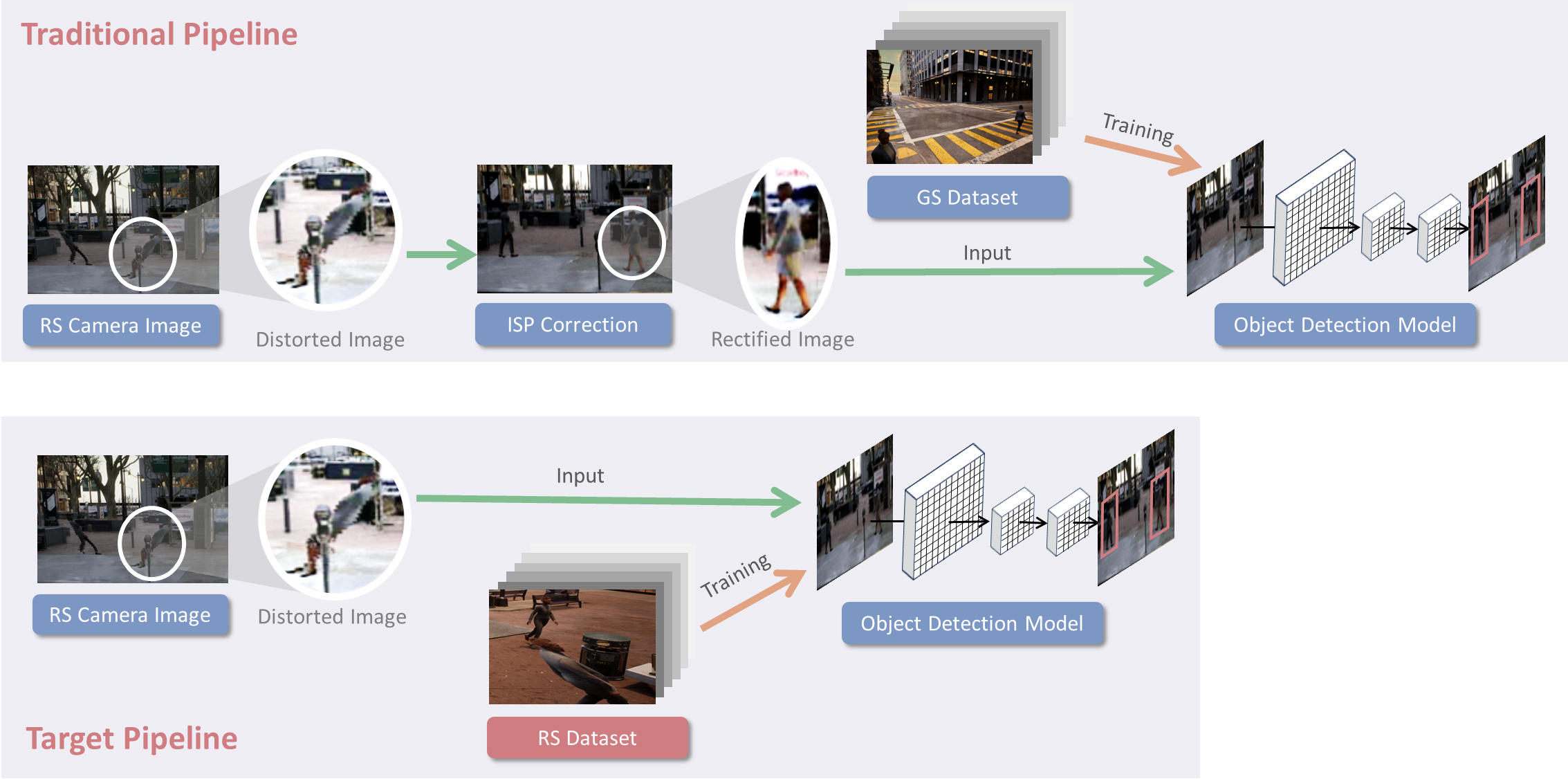}
    \caption{Comparison of the baseline and our target CV pipelines; the latter avoids ISP correction for RS images.}
    \label{fig:target}
\end{figure}

Meanwhile, efforts to mitigate the energy consumption of camera-driven CV pipelines have gained a lot of attention in the literature, particularly for energy-constrained applications, such as autonomous drones, surveillance, and headsets for augmented reality~\cite{bian2023colibriuav,surveillance}. The research advocates bringing the compute-heavy ML algorithms as close to the image sensor as possible \cite{chen2020pns,sony2020vision}. This co-location can reduce the energy associated with transferring large amounts of sensor data between chips and, when taken to the extreme of implementing in-sensor computing, minimize the cost of energy-expensive analog-to-digital conversion within the sensor \cite{datta2022scireports}. Unfortunately, the complexity of many algorithms, including RS correction, makes them difficult to implement in and near the sensor. This presents a compelling question: is the RS correction in a ML pipeline necessary? Instead, can the ML pipeline automatically compensate for RS artifacts as shown in Fig.~\ref{fig:target}? 

To evaluate this question, we propose to use a rolling shutter (RS) dataset for training and fine-tuning ML models and compare their accuracies with those trained with a global shutter (GS) dataset. Unfortunately, existing public datasets tailored for studying the RS effect lack this crucial pairing with a GS dataset~\cite{schubert2019vidsors,9027901,cao2022learning}. This void in resources compelled us to leverage the CV simulation capabilities of Unreal Engine 5 (UE5). We first generated a GS dataset with a very high frame rate. To create the RS 
dataset, we then emulated the rolling shutter effect by amalgamating successive rows 
from sequences of generated GS images, mirroring the characteristic line-by-line scan 
intrinsic to rolling shutters.
\begin{figure}[!t]
     \centering
     \begin{subfigure}[b]{0.6\linewidth}
         \centering
         \includegraphics[width=\linewidth]{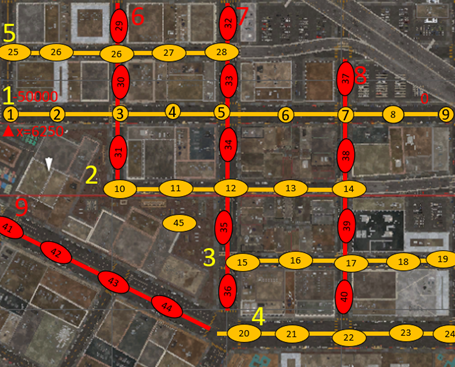}
         \caption{Scenes map}
         \label{fig:map}
     \end{subfigure} \\
     \begin{subfigure}[b]{0.9\linewidth}
         \centering
         \includegraphics[width=\linewidth]{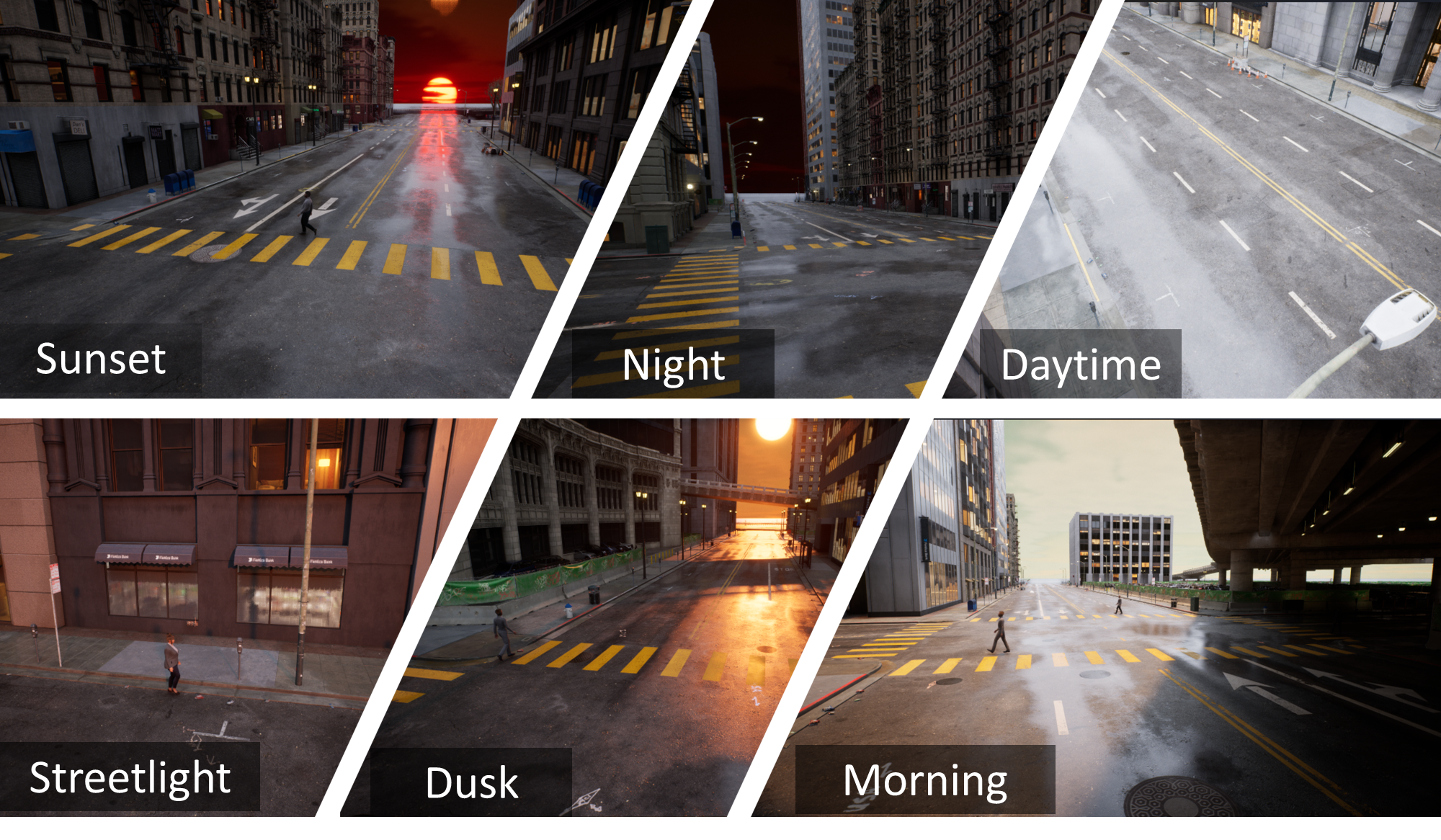}
         \caption{Varying time of day}
         \label{fig:scenes}
     \end{subfigure}
        \caption{Unreal Engine 5 Dataset Scene Generation}
        \label{fig:dataset}
\end{figure}
This work makes three key contributions to the field of object detection under different camera shutter effects.
\vspace*{-0.25em}
\begin{itemize}
\itemsep-0.25em 
\item \textbf{Synthetic Dataset Generation for Shutter Effect Analysis:} We generate a synthetic paired GS/RS dataset using the real-time 3D creation software Unreal Engine 5 (UE5) designed to evaluate pedestrian detection models for both rolling and global shutters under various conditions, 
as illustrated in Fig.~\ref{fig:dataset}.
\item \textbf{Empirical Validation of Detection Models under Different Shutter Effects:} We use synthetic dataset to conduct extensive experiments on mainstream object detection models, specifically YOLOv8 and DETR, to assess their performance under different shutter effects. The results show that ML pipelines need not correct for a RS for many coarse-grained object detection applications. However, for applications that require fine-grained location of the objects, the results suggest that achieving ISP-less CV pipelines for RS cameras may need additional effort. Our results also show that the accuracy of the pedestrian detection models can be significantly improved with our synthetic dataset while retaining their transferability to GS images.
\item \textbf{Development of a Shutter Simulation Framework:} We have developed a comprehensive framework that simulates ultra-high frame rate GS images in order to simulate RS effects, providing a versatile toolset for generating pedestrian detection datasets under various shutter conditions. 
\end{itemize}

\section{Rolling Shutter vs Global Shutter}

\begin{figure}[!t]
     \centering
     \begin{subfigure}[b]{0.45\linewidth}
         \centering
         \includegraphics[width=\linewidth]{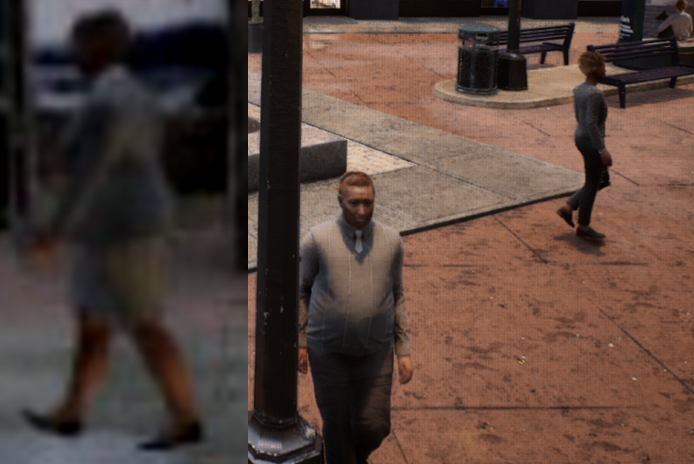}
         \caption{Global Shutter}
         \label{fig:gs}
     \end{subfigure}
     \begin{subfigure}[b]{0.45\linewidth}
         \centering
         \includegraphics[width=\linewidth]{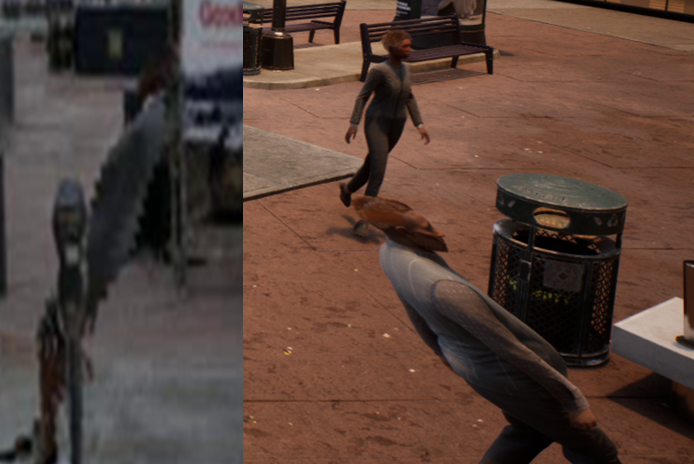}
         \caption{Rolling Shutter}
         \label{fig:rs}
     \end{subfigure}
        \caption{Rolling Shutter VS Global Shutter}
        \label{fig:examole}
\end{figure}

A digital camera typically captures images using either a RS or a GS mechanism. The primary distinction between the two is in the way they capture and process light onto the sensor. GS captures the entire image simultaneously. Every pixel on the sensor is exposed to light at the exact same moment, resulting in a distortion-free capture of fast-moving subjects, as shown in Fig.~\ref{fig:gs}.  In the RS mode, the image sensor scans and captures the image line by line, sequentially from top to bottom. This means that not all parts of the image are recorded at precisely the same time. For subjects in fast motion or when the camera itself is moving quickly, this sequential capturing can result in distortions, commonly referred to as `rolling shutter artifacts' or the `jello effect'. An example artifact is when pedestrians look bent or skewed, as shown in Fig.~\ref{fig:rs}.

From the perspective of peripheral circuit design, the mechanism of choice has implications for readout circuitry, speed, and complexity. The choice is often a trade-off between cost, speed, and potential image artifacts~\cite{kandula2020deep,huai2023automated,koh2011novel}.

When using RS cameras for CV applications, the detected position of the object can be affected by the time delay between the top and bottom rows of the sensor. This can result in misalignment or incorrect positioning of the detected object in the processed image. For instance, a fast-moving car could appear slightly tilted or elongated when captured by a camera with a RS mechanism, potentially leading to less accurate detection or misinterpretation of its speed and trajectory. We will investigate the impact of RS images on the detection performance of object detection models in the following sections.

\vspace{-0.4cm}
\section{Analyzing the Impact of RS using UE5}

Capitalizing on UE5's capabilities, we used the "sample city" project~\cite{UnrealEngine2023} 
as our foundational city environment to conduct our experiment. 
We designed and implemented 40 distinct urban street scenes that span across 8 streets as shown in Fig.~\ref{fig:map}. Each of these scenes showcases a unique environment adding diversity to the dataset.

\textbf{Temporal Setting Variations:} Every individual street scene is rendered under five different times of the day, emulating a comprehensive spectrum of lighting conditions. These distinct times are visually represented by varying light intensities and angles, as depicted in Fig.~\ref{fig:scenes}.

\textbf{Crowd Dynamics:} The scenes incorporate randomized crowds to mimic real-world scenarios. Factors such as gender, height, body shape, skin tone, hair, and attire vary to introduce diversity and realism. Considering that pedestrians play a vital role in our analysis, their maximum walking speed is 2 meters per second in our normal walking speed dataset. 
Additionally, we provide a dataset with the pedestrians walking at 10x normal speed to study the detection model's performance on RS images under faster motion conditions.

\textbf{Camera Settings and Global Shutter Data Generation:}
To capture the nuances of each scene and the effect of RS on object detection, we use the following camera settings: an aperture of f/2.8, a focal length of 35.0mm, a filmback ratio of 16:9 for digital film, a 12mm prime lens at f/2.8, and a frame rate set at 32,400fps.

Each scene was documented using five cameras positioned at diverse angles, with each camera continuously capturing 1080 frames per camera for each environmental condition. All images maintain a resolution of $1920\times1080$. Moreover, for every pedestrian that made an appearance in a given shot, we generated a bounding box annotation. 

We use the first frame out of the 1080 frames as a frame in the GS dataset where an entire sequence of 1080 frames are used to create a single frame in the RS dataset, as described below.  Thus, the GS dataset has a frame rate of 30 frames per second (i.e., $1080 \times 30 = 32400$), a typical rate for cameras.

\textbf{Generation of Rolling Shutter Dataset:}
To synthesize the RS dataset, we simulate the RS effect by sequentially replacing rows of pixels in a top-to-bottom fashion with the corresponding rows of a sequence of GS images, emulating the line-by-line scan typical of rolling shutters. Thus, for each sequence of 1080 images from the GS dataset, we produce a single image that captures the RS effect. 

Following the generation of the RS images, we use an annotation tool~\cite{AutoLabelImg_yolov7} to manually label each pedestrian present in every frame, providing the ground truth needed to train our object detection models. 


\section{Experimental Results}

\textbf{Dataset Specifications and Distribution:}
In this paper, we generate four distinct datasets, namely Normal\_RS and Normal\_GS for pedestrians walking at a normal walking speed of $2m/s$ and Faster\_RS and Faster\_GS for pedestrians moving $10\times$ faster than normal walking speed. 
These datasets all have 1000 frames, where 800 for training, 100 for validation, and 100 for testing. 
For the Normal\_RS training dataset, the average size of a bounding box is $7,725 \text{px}^2$ with a total of $2,428$ bounding boxes, yielding an average of $3.17$ bounding boxes per image. 
Similarly, the Normal\_GS training dataset has an average bounding box size of $10,847 \text{px}^2$ and a cumulative count of $2,337$ bounding boxes with an average of $3.18$ bounding boxes across the images containing them. 
The Faster\_RS training dataset has an average bounding box size of $11,465 \text{px}^2$, with $2,560$ bounding boxes in total and $3.31$ boxes per image on average. 
Lastly, the Faster\_GS training dataset has an average bounding box size of $11,932 \text{px}^2$. 
The total number of bounding boxes is $2,616$ with an average of $3.43$ boxes per image. 

The minor differences between the bounding box sizes of the GS and RS datasets can be attributed to differences between manual labelling of the RS datasets and the automatic labelling (by UE5) of the GS dataset. The minor difference between the number of frames can be due to the fact that GS 
datasets sample the image at the beginning of the frame interval, whereas RS takes images across the entire frame interval.

\begin{figure}[!t]
    \centering
    \includegraphics[width=0.9\linewidth]{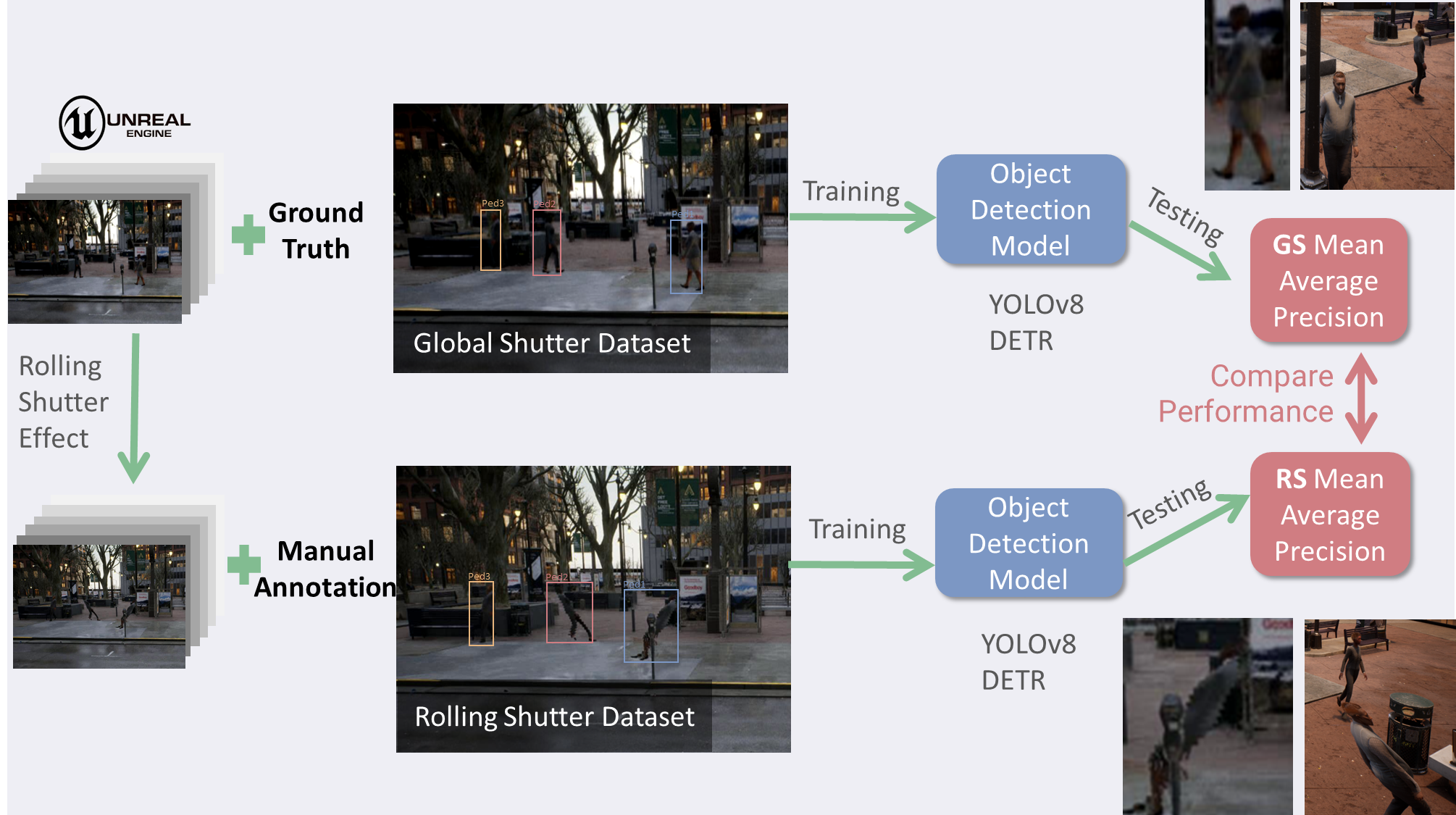}
    \caption{Pedestrians Detection Experiment Pipeline}
    \label{fig:pipeline}
\end{figure}

\textbf{Performance of RS and GS Datasets:}
%
For the object detection efficacy assessment on these datasets, we tested datasets on the state-of-the-art model YOLOv8~\cite{Jocher_YOLO_by_Ultralytics_2023} and the transformer-based model DETR~\cite{carion2020end}. Training for YOLOv8 was conducted with a set learning rate of 0.001, while DETR utilized a learning rate of 0.0001, with both models being trained for 100 epochs. The procedural flow of our experiment is graphically represented in Fig.~\ref{fig:pipeline}. The performance metrics that we evaluated are precision (P), recall (R), and mean average precision (mAP) measured with Intersection over Union (IoU) thresholds of 0.5 as well as from a range of 0.5 to 0.95 with a step size of 0.05\footnote{Note that DETR model does not provide a metric for precision.}.

We pretrained all of our models on the COCO dataset and then validated the normal and faster datasets with YOLOv8 and DETR.
%
%
The results presented in Table~\ref{Tab:results} reveal that the mean Average Precision (mAP) results for training and validation on GS and RS, when the Intersection Over Union (IOU) threshold is set at 0.5, are remarkably similar, with ${<}2\%$ deviation. Remarkably, this congruence holds for both slower and faster walking conditions. However, when the detection criteria is more stringent, measured with an IOU that ranges from 0.5 to 0.95, the discrepancy grows to 24\%. 

These results suggest that for coarse-grained detection of objects, it may not be necessary to 
correct the RS effect in CV pipeline.
However, the results also show that the distortion of the pedestrians makes accurate sizing and positioning of their bounding boxes more difficult, and that this uncertainty is difficult to compensate with training.\footnote{Note that the faster datasets yield, on average, higher mAP scores than the normal dataset. We conjecture that this is due to the variation in the scene generation which resulted in the faster datasets having pedestrians that are closer to the camera and thus seem larger.}

\begin{table}[!t]
\centering
\caption{Model Performance with GS and RS Datasets}
\label{Tab:results}
\resizebox{\linewidth}{!}{
\begin{tabular}{ccccccccc}
\toprule
\multirow{2}{*}{Model} & \multirow{2}{*}{Dataset}&
\multirow{2}{*}{P} & \multirow{2}{*}{R} & \multicolumn{2}{c}{mAP}   \\ 
\cline{5-6}
 &  & &    & IOU@0.5 & IOU@0.5:0.95 \\
\midrule
\multirow{2}{*}{YOLOV8} & Normal\_GS&  0.97 & 0.70 & 0.82 & 0.60\\
                        & Normal\_RS&  0.94 & 0.67 & 0.82 & 0.44 \\
\midrule
\multirow{2}{*}{DETR} & Normal\_GS&  $-$ & 0.51 &	0.72 &	0.40\\
                      & Normal\_RS&  $-$ & 0.40	& 0.71	& 0.28\\

\midrule
\multirow{2}{*}{YOLOV8} & Faster\_GS&  0.99 & 0.97 & 0.98 & 0.72\\
                        & Faster\_RS&  0.98 & 0.97 & 0.99 & 0.59 \\
\midrule
\multirow{2}{*}{DETR} & Faster\_GS&  $-$ &0.64	& 0.96	& 0.53\\
                      & Faster\_RS&  $-$ & 0.61	& 0.98 &	0.48\\
\bottomrule
\end{tabular}}
\end{table}

\begin{table}[!t]
\centering
\caption{Accuracy Drop when Trained only on COCO}
\label{Tab:COCOpretrained}
\resizebox{\linewidth}{!}{
\begin{tabular}{cccccccc}
\toprule
\multirow{2}{*}{Validation}&
\multirow{2}{*}{P} & \multirow{2}{*}{R} & \multicolumn{2}{c}{mAP}   \\ 
\cline{4-5}
 &  &  & IOU@0.5 & IOU@0.5:0.95 \\
\midrule
Normal\_GS & 0.35  &    0.21   &   0.12  &   0.02\\
Normal\_RS & 0.91   &   0.55   &   0.67  &    0.31 \\
\midrule
Faster\_GS &  0.23  &    0.16  &    0.05  &  0\\
Faster\_RS & 0.13   &   0.10    &  0.05  &  0 \\
\bottomrule
\end{tabular}}
\vspace*{-0.5cm}
\end{table}

\textbf{Cross-Training and Validation of Datasets:}
We also measured the detection outcomes of the models that are trained exclusively on COCO using YOLOv8. Comparing the results in Table~\ref{Tab:COCOpretrained} with that of Table~\ref{Tab:results}, we see that training only on COCO yields significantly worse results for both mAP with IOU@0.5 and IOU@0.5:0.95. This shows the importance of fine-tuning these models on application-specific 
datasets and, in particular, shows the value of our datasets for pedestrian detection.

The second dataset analysis shown in Table~\ref{Tab:crossval} presents the results on YOLOv8 with a combination of fast and slow pedestrians. The results show that training on our RS dataset significantly improves the test mAP of RS images compared to training on GS images, showing the efficacy of our dataset.  Moreover, models trained with the RS dataset perform similarly when tested on a combination of RS and GS images, showing the transferability of the models to GS images.

\begin{table}[!t]
\centering
\caption{RS \& RS+GS Dataset Validation on YOLOv8 with Fast and Slow Pedestrians}
\label{Tab:crossval}
\resizebox{\linewidth}{!}{
\begin{tabular}{ccccccc}
\toprule
 \multirow{2}{*}{Train} & \multirow{2}{*}{Validation} & 
\multirow{2}{*}{P} & \multirow{2}{*}{R} & \multicolumn{2}{c}{mAP} &  \\ 
\cline{5-6}
 &  &  &  & IOU@0.5 & IOU@0.5:0.95 \\
\midrule
         RS & RS &  0.90  &    0.75  &    0.82  &     0.35\\
         GS & RS & 0.80    &  0.67    &  0.63    &  0.23 \\   
         RS & GS+RS &  0.94  &    0.71   &   0.80  &    0.36 \\
\bottomrule
\end{tabular}}
\end{table}

\begin{figure}[!t]
     \centering
     \begin{subfigure}[b]{0.49\linewidth}
         \centering
         \includegraphics[width=\linewidth]{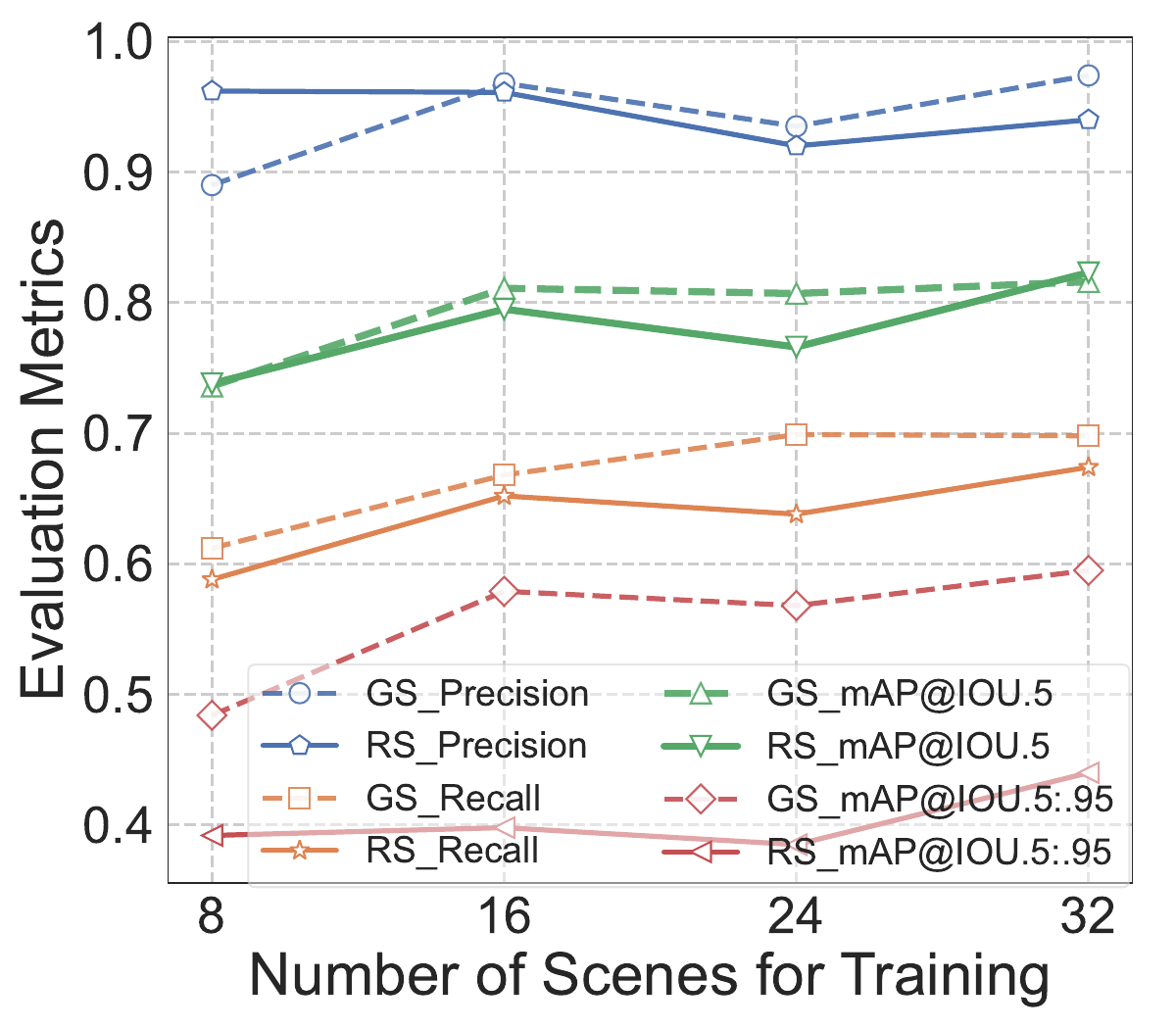}
         \caption{Impact of Scenes}
         \label{fig:reducetrain}
     \end{subfigure}
     \begin{subfigure}[b]{0.49\linewidth}
         \centering
         \includegraphics[width=\linewidth]{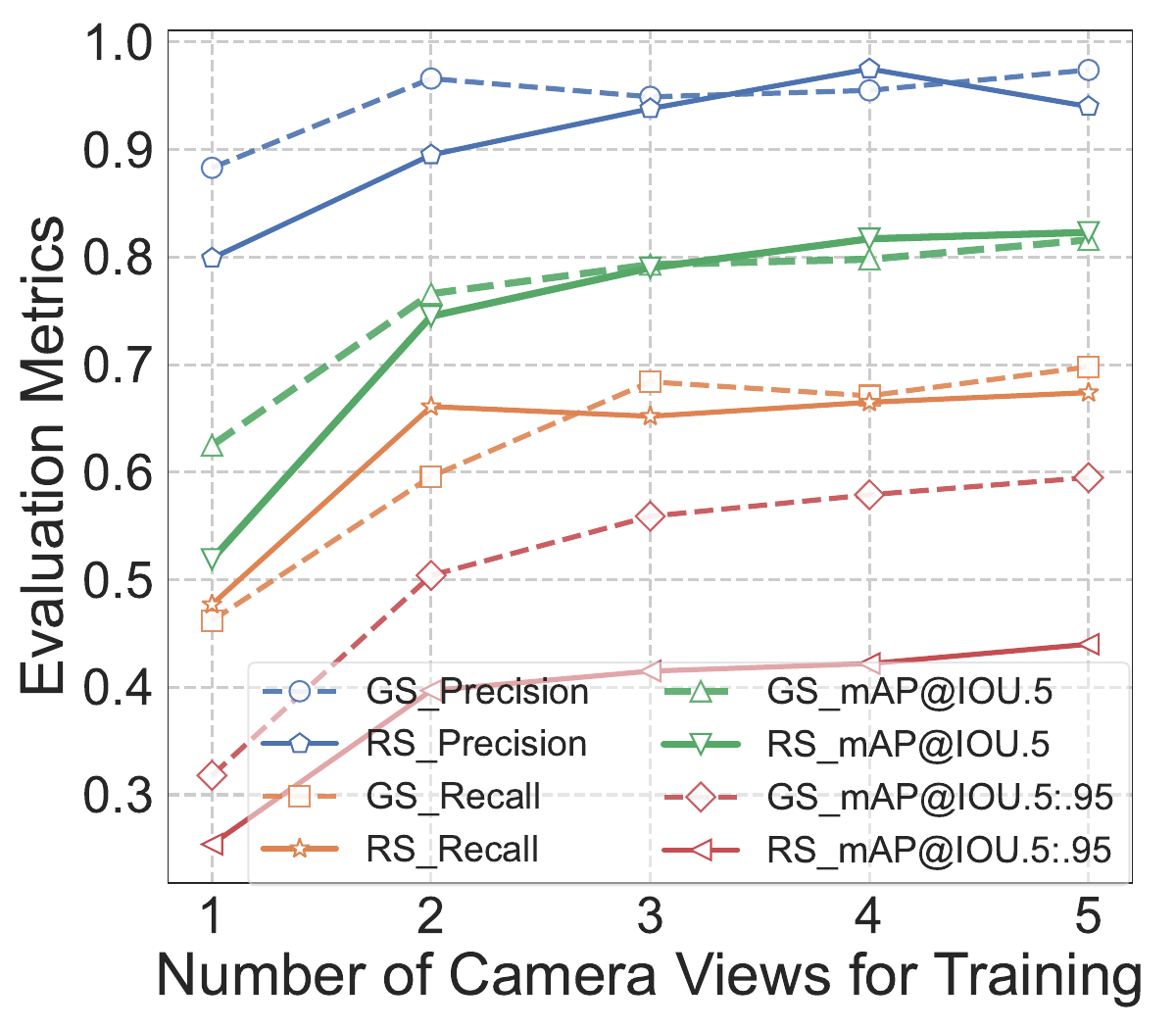}
         \caption{Impact of Camera Views}
         \label{fig:camangle}
     \end{subfigure}
        \caption{Impact of the pedestrian detection performance on the deiversity of our dataset}
        \vspace{-3mm}
        \label{fig:diversity}
\end{figure}

Lastly, in order to show the evaluation results with different diversity, we analyze the impact of the number of scenes and camera views in the training dataset in Fig.~\ref{fig:diversity}. It shows that, with the increasing of training dataset size, which also increases the scene diversity in Fig.~\ref{fig:reducetrain} and camera views diversity in Fig.~\ref{fig:camangle}, the mAP generally increases. 

\vspace{-2mm}
\section{Conclusions}

This paper analyzes the intricate relationship between camera shutter mechanisms and their implications on pedestrian detection models. Our findings identify the relative degree of accuracy achievable in ML detection accuracy between global and RS modalities. In particular, they show that RS correction is not necessary in scenarios where moderately-grained overlap of the bounding boxes with the ground truth is necessary, i.e., using the mean average precision metric with an IoU of 0.5. 
This result challenges the prevailing notion that RS corrections are indispensable for 
all camera operations, suggesting that for specific applications like pedestrian detection, such corrections might be unnecessary~\cite{dai2016rolling,schubert2019rolling,lee2017inertial,mo2022imu}. This results helps quantify the impact of RS effects in recently proposed energy-efficient smart camera systems that propose to limit the application of an ISP pipeline and leverage in-pixel computing paradigms~\cite{datta2022eccv,datta2022p2mdetrack,datta2023icassp}. 

Our work's significance is amplified by the introduction of a synthetic dataset, hand-crafted using Unreal Engine 5 (UE5). This dataset, simulating ultra-high frame-rate GS images to emulate 
the impact of RS, stands as a testament to the fusion of advanced simulation capabilities 
with practical CV applications for RS cameras. 



\small{
\bibliographystyle{IEEEbib}
\bibliography{refs}

\begin{thebibliography}{10}

\bibitem{dai2016rolling}
Y.~Dai, H.~Li, and L.~Kneip,
\newblock ``Rolling shutter camera relative pose: Generalized epipolar
  geometry,''
\newblock in {\em Proceedings of the IEEE Conference on Computer Vision and
  Pattern Recognition}, 2016, pp. 4132--4140.

\bibitem{fan2021sunet}
B.~Fan, Y.~Dai, and M.~He,
\newblock ``{SUNet}: symmetric undistortion network for rolling shutter
  correction,''
\newblock in {\em Proceedings of the IEEE/CVF International Conference on
  Computer Vision}, 2021, pp. 4541--4550.

\bibitem{lao2018robust}
Y.~Lao and O.~Ait-Aider,
\newblock ``A robust method for strong rolling shutter effects correction using
  lines with automatic feature selection,''
\newblock in {\em Proceedings of the IEEE Conference on Computer Vision and
  Pattern Recognition}, 2018, pp. 4795--4803.

\bibitem{rengarajan2017unrolling}
V.~Rengarajan, Y.~Balaji, and A.~Rajagopalan,
\newblock ``Unrolling the shutter: {CNN} to correct motion distortions,''
\newblock in {\em Proceedings of the IEEE Conference on computer Vision and
  Pattern Recognition}, 2017, pp. 2291--2299.

\bibitem{zhuang2019learning}
B.~Zhuang, Q.-H. Tran, P.~Ji, L.-F. Cheong, and M.~Chandraker,
\newblock ``Learning structure-and-motion-aware rolling shutter correction,''
\newblock in {\em Proceedings of the IEEE/CVF Conference on Computer Vision and
  Pattern Recognition}, 2019, pp. 4551--4560.

\bibitem{liu2020deep}
P.~Liu, Z.~Cui, V.~Larsson, and M.~Pollefeys,
\newblock ``Deep shutter unrolling network,''
\newblock in {\em Proceedings of the IEEE/CVF Conference on Computer Vision and
  Pattern Recognition}, 2020, pp. 5941--5949.

\bibitem{zhuang2017rolling}
B.~Zhuang, L.-F. Cheong, and G.~Hee~Lee,
\newblock ``Rolling-shutter-aware differential {SfM} and image rectification,''
\newblock in {\em Proceedings of the IEEE International Conference on Computer
  Vision}, 2017, pp. 948--956.

\bibitem{cao2022learning}
M.~Cao, Z.~Zhong, J.~Wang, Y.~Zheng, and Y.~Yang,
\newblock ``Learning adaptive warping for real-world rolling shutter
  correction,'' 2022.

\bibitem{bian2023colibriuav}
S.~Bian, L.~Schulthess, G.~Rutishauser, A.~Di Mauro, L.~Benini, and M.~Magno,
\newblock ``{ColibriUAV}: An ultra-fast, energy-efficient neuromorphic edge
  processing {UAV}-platform with event-based and frame-based cameras,'' 2023.

\bibitem{surveillance}
J.~Xie, Y.~Zheng, R.~Du, W.~Xiong, Y.~Cao, Z.~Ma, D.~Cao, and J.~Guo,
\newblock ``Deep learning-based computer vision for surveillance in its:
  Evaluation of state-of-the-art methods,''
\newblock {\em IEEE Transactions on Vehicular Technology}, vol. 70, no. 4, pp.
  3027--3042, 2021.

\bibitem{chen2020pns}
Z.~Chen, H.~Zhu, E.~Ren, Z.~Liu, K.~Jia, L.~Luo, X.~Zhang, Q.~Wei, F.~Qiao,
  X.~Liu, and H.~Yang,
\newblock ``Processing near sensor architecture in mixed-signal domain with
  {CMOS} image sensor of convolutional-kernel-readout method,''
\newblock {\em IEEE Transactions on Circuits and Systems I: Regular Papers},
  vol. 67, no. 2, pp. 389--400, 2020.

\bibitem{sony2020vision}
``{Sony to Release World's First Intelligent Vision Sensors with {AI}
  Processing Functionality},''
  \url{https://www.sony.com/en/SonyInfo/News/Press/202005/20-037E/}, 2020,
\newblock Accessed: 12-01-2022.

\bibitem{datta2022scireports}
G.~Datta et~al.,
\newblock ``A processing-in-pixel-in-memory paradigm for resource-constrained
  {TinyML} applications,''
\newblock {\em Scientific Reports}, vol. 12, 2022.

\bibitem{schubert2019vidsors}
D.~Schubert, N.~Demmel, L.~von Stumberg, V.~Usenko, and D.~Cremers,
\newblock ``Rolling-shutter modelling for visual-inertial odometry,''
\newblock in {\em International Conference on Intelligent Robots and Systems
  (IROS)}, November 2019.

\bibitem{9027901}
Like Cao, Jie Ling, and Xiaohui Xiao,
\newblock ``The whu rolling shutter visual-inertial dataset,''
\newblock {\em IEEE Access}, vol. 8, pp. 50771--50779, 2020.

\bibitem{kandula2020deep}
P.~Kandula, T.~L. Kumar, and A.N. Rajagopalan,
\newblock ``Deep end-to-end rolling shutter rectification,''
\newblock {\em JOSA A}, vol. 37, no. 10, pp. 1574--1582, 2020.

\bibitem{huai2023automated}
J.~Huai, Y.~Zhuang, B.~Wang, C.~Zhang, Y.~Shao, J.~Tang, and A.~Yilmaz,
\newblock ``Automated rolling shutter calibration with an {LED} panel,''
\newblock {\em Optics Letters}, vol. 48, no. 4, pp. 847--850, 2023.

\bibitem{koh2011novel}
I.-S. Koh, S.-Y. Ro, J.-P. Kim, K.-Y. Min, and J.-W. Chong,
\newblock ``A novel digital image stabilization for mobile applications,''
\newblock in {\em 2011 IEEE International Conference on Consumer Electronics
  (ICCE)}. IEEE, 2011, pp. 209--210.

\bibitem{UnrealEngine2023}
``City sample project unreal engine demonstration,'' Unreal Engine
  Documentation, 2023,
\newblock Available at:
  \href{https://docs.unrealengine.com/5.1/en-US/city-sample-project-unreal-engine-demonstration/}{https://docs.unrealengine.com/5.1/en-US/city-sample-project-unreal-engine-}\\
  \href{https://docs.unrealengine.com/5.1/en-US/city-sample-project-unreal-engine-demonstration/}{demonstration/}.

\bibitem{AutoLabelImg_yolov7}
ZayneYe,
\newblock ``{AutoLabelImg}: Multifunction autoannotate tools,'' GitHub
  repository, 2022,
\newblock Available at: \url{https://github.com/ZayneYe/AutoLabelImg_yolov7}.

\bibitem{Jocher_YOLO_by_Ultralytics_2023}
G.~Jocher, A.~Chaurasia, and J.~Qiu,
\newblock ``{YOLO by Ultralytics},'' Jan. 2023.

\bibitem{carion2020end}
N.~Carion, F.~Massa, G.~Synnaeve, N.~Usunier, A.~Kirillov, and S.~Zagoruyko,
\newblock ``End-to-end object detection with transformers,''
\newblock in {\em European conference on computer vision}. Springer, 2020, pp.
  213--229.

\bibitem{schubert2019rolling}
D.~Schubert, N.~Demmel, L.~von Stumberg, V.~Usenko, and D.~Cremers,
\newblock ``Rolling-shutter modelling for direct visual-inertial odometry,''
\newblock in {\em 2019 IEEE/RSJ International Conference on Intelligent Robots
  and Systems (IROS)}. IEEE, 2019, pp. 2462--2469.

\bibitem{lee2017inertial}
Chang-Ryeol Lee and Kuk-Jin Yoon,
\newblock ``Inertial-aided rolling shutter relative pose estimation,''
\newblock {\em arXiv preprint arXiv:1712.00184}, 2017.

\bibitem{mo2022imu}
J.~Mo, M.~J. Islam, and J.~Sattar,
\newblock ``{IMU}-assisted learning of single-view rolling shutter
  correction,''
\newblock in {\em Conference on Robot Learning}. PMLR, 2022, pp. 861--870.

\bibitem{datta2022eccv}
G.~Datta, Z.~Yin, A.~P. Jacob, A.~R. Jaiswal, and P.~A. Beerel,
\newblock ``Towards energy-efficient hyperspectral image processing inside
  camera pixels,''
\newblock in {\em Computer Vision -- ECCV 2022 Workshops}. 2023, pp. 303--316,
  Springer.

\bibitem{datta2022p2mdetrack}
G.~Datta, S.~Kundu, Z.~Yin, J.~Mathai, Z.~Liu, Z.~Wang, M.~Tian, S.~Lu, R.~T.
  Lakkireddy, A.~Schmidt, W.~Abd-Almageed, A.~Jacob, A.~Jaiswal, and P.~Beerel,
\newblock ``{P2M-DeTrack}: Processing-in-pixel-in-memory for energy-efficient
  and real-time multi-object detection and tracking,''
\newblock in {\em VLSI-SoC}, 2022, pp. 1--6.

\bibitem{datta2023icassp}
G.~Datta, Z.~Liu, M.~A. Kaiser, S.~Kundu, J.~Mathai, Z.~Yin, A.~P. Jacob, A.~R.
  Jaiswal, and P.~A. Beerel,
\newblock ``In-sensor \& neuromorphic computing are all you need for energy
  efficient computer vision,''
\newblock in {\em ICASSP}, 2023, pp. 1--5.

\end{thebibliography}
}

\end{document}